\documentclass[lettersize,journal]{IEEEtran}

\usepackage{amsmath,amssymb,mathtools}
\usepackage{graphicx,booktabs,multirow}
\usepackage{algorithm,algpseudocode}
\usepackage{xcolor}
\usepackage{enumitem,cite}

\usepackage{subcaption}

\newcommand{\mc}{\mathcal}
\newcommand{\bb}{\mathbb}

\usepackage{titlesec}

\titleformat{\paragraph}[runin]
  {\normalfont\itshape}
  {}
  {0pt}
  {}

\titlespacing*{\paragraph}
  {0pt}  
  {0.6\baselineskip}
  {0.5em}       

\def \mc {\mathcal}

\title{Structured LLM Reasoning for Zero-Shot Human--Robot Coordination Under Hidden Goals
\thanks{This work was supported in part  by the ONR
award N00014-22-1-2813.}
}

\author{Dong Hae Mangalindan, Anand Gokhale, Francesco Bullo, and Vaibhav Srivastava
\thanks{D. Mangalindan and V. Srivastava are with the Department of Electrical and Computer Engineering, Michigan State University, East Lansing, MI 48824.  e-mail: \texttt{\small \{mangalin, vaibhav\}@msu.edu}}
\thanks{A. Gokhale and F. Bullo are  with
    the Center for Control, Dynamical Systems, and Computation, UC Santa
    Barbara, Santa Barbara, CA 93106 USA.  {\tt\small
      \{anand\_gokhale,bullo\}@ucsb.edu}}
}

\usepackage{tikz}
\usetikzlibrary{
    arrows.meta,
    positioning,
    calc,
    fit
}

\tikzset{
    >={Latex[length=1.8mm,width=1.3mm]},
    block/.style={
        draw=black,
        rectangle,
        rounded corners=1pt,
        line width=0.55pt,
        fill=white,
        align=center,
        font=\footnotesize,
        inner xsep=4pt,
        inner ysep=3pt,
        minimum height=7mm
    },
    llmblock/.style={
        block,
        rounded corners=2pt,
        fill=black!8,
        line width=0.75pt
    },
    exactblock/.style={
        block,
        fill=black!2,
        double=black,
        double distance=0.55pt,
        line width=0.4pt
    },
    arr/.style={
        draw=black,
        line width=0.65pt,
        -{Latex[length=1.8mm,width=1.3mm]}
    },
    darr/.style={
        arr,
        dashed,
        dash pattern=on 2.4pt off 1.6pt
    },
    flowlabel/.style={
        font=\scriptsize,
        align=center,
        inner sep=1pt,
        fill=white
    }
}

\begin{document}
\maketitle

\begin{abstract}
We present a structured large-language-model (LLM) architecture for zero-shot human--robot coordination in a cooperative construction task with private goal views. Guided by a Dec-POMDP formulation, the architecture decomposes decision-making into (i) action-conditioned Theory-of-Mind (ToM) inference, (ii) hierarchical planning, (iii) conversation interpretation, (iv) action verification, and (v) feedback-based replanning. We compare the proposed method with an ablation without ToM inference and a multi-agent reinforcement-learning policy trained offline over many goal pairs. In human-participant experiments, the proposed method required fewer interaction steps and yielded higher post-interaction trust ratings than both baselines. These results suggest that systematically decomposing the team decision problem, using LLMs as tractable surrogates for otherwise intractable inference and planning computations, and retaining conventional verification for physical feasibility can improve both task coordination and the human experience.
\end{abstract}


\section{Introduction}

Autonomous systems increasingly function as members of mixed human-robot teams rather than as isolated decision makers. Such teams often operate under asymmetric information: each teammate may possess a private goal, observation, or preference that is not directly available to the other. Successful coordination therefore requires the robot to infer task-relevant hidden variables from interaction history, communicate when needed, and revise its plan as new evidence becomes available, while respecting exact constraints imposed by geometry, dynamics, and available resources. Decentralized partially observable Markov decision processes (Dec-POMDPs) provide a formal framework for such problems, but exact inference and planning quickly become computationally intractable. The growing reasoning capabilities of LLMs raise the possibility of preserving this decision-theoretic structure while using LLMs as tractable surrogates for the intractable computations and retaining conventional mechanisms where exact verification is required.

We study this problem in a cooperative construction task inspired by the La Boca game. A human and a robot jointly build a three-dimensional structure, each given only a private two-dimensional target projection. Because neither view alone specifies the desired structure, the robot must coordinate under uncertainty about the human’s goal while ensuring that each proposed action remains physically feasible. We develop a structured LLM architecture guided by the underlying team-decision problem and show that, in our experiments, it improves coordination efficiency and human trust relative to the considered baselines.

LLMs have been used for high-level planning, task decomposition, code
generation, and language-conditioned decision making, including grounding
semantic decisions in executable skills and revising plans using execution
feedback~\cite{ahn2022can,huang2022inner,liang2023code,
singh2022progprompt,huang2022language,wang2023voyager}. More structured
architectures organize memory, reasoning, action selection, and
recovery~\cite{sumers2023cognitive}, invoke recursive decomposition following
execution failure~\cite{prasad2024adapt}, or combine language-based reasoning
with persistent knowledge graphs and symbolic
planning~\cite{shek2026kglamp}. In multi-agent systems, LLMs have been used as
coordinators, communicators, and memory modules~\cite{li2025language}, to
support action-by-action embodied collaboration~\cite{white2025collaborating},
and to learn human-interpretable communication jointly with cooperative
policies~\cite{li2024language}. Related HRI architectures combine LLM planning
with multimodal observations and physical and communicative
actions~\cite{wang2024lami}.

Complementary safety-oriented work places external safeguards around
LLM-generated plans. These approaches use reachability analysis,
temporal-logic constraints, or programmatic verification to validate or modify
plans before execution~\cite{hafez2025safe,ravichandran2026safety,
gokhale2025logicguard}. Together, this literature supports hybrid systems in
which LLMs provide semantic reasoning and long-horizon planning, while
conventional modules provide grounding, execution, and verification.

A parallel body of work studies intent inference, inverse planning, and
computational theory of mind. These methods interpret observed actions as
evidence about latent goals, beliefs, or preferences~\cite{baker2009action,gmytrasiewicz2005framework,
albrecht2018autonomous}. Related HRI work applies these ideas to legible robot
behavior and human-intent inference~\cite{dragan2013legibility}.

Dec-POMDPs provide a formal model for cooperative decision making under
private information and partial observability~\cite{bernstein2002complexity,oliehoek2016concise}, although exact solution
methods become computationally intractable beyond small problems. Multi-agent
reinforcement learning provides a complementary computational paradigm in
which coordination policies are learned through repeated interaction~\cite{lowe2017multi,rashid2018qmix,yu2022surprising}.

Building on these foundations, we develop a structured LLM architecture guided
by a Dec-POMDP formulation of the cooperative task. Specifically, we use LLMs as practical surrogates for otherwise intractable inference and planning
computations, while retaining conventional verification to enforce physical feasibility. The resulting architecture decomposes decision making into ToM inference, hierarchical planning, conversation interpretation, action
verification, and feedback-based replanning. Whereas prior approaches discussed above typically focus on only a subset of these capabilities, the proposed architecture integrates them within a unified framework for human-robot coordination under private information.

The major contributions of this work are twofold. First, we develop a
Dec-POMDP-guided structured LLM architecture for human-robot coordination
under private information, combining approximate inference and planning with
rule-based feasibility verification. Second, we evaluate the approach against
an LLM ablation and an offline-trained multi-agent RL baseline in
human-participant experiments, considering coordination efficiency, task
completion, and perceived trust.

The remainder of the paper is organized as follows.
Section~\ref{sec:problem} introduces the cooperative construction task and its
Dec-POMDP formulation. Section~\ref{sec:architecture} presents the structured
LLM coordination architecture. Section~\ref{sec:experiments} describes the
offline RL baseline, human-participant experiment, and experimental results.
Section~\ref{sec:discussion} discusses the findings, limitations, and future
directions.

\section{Cooperative Construction as a Dec-POMDP}
\label{sec:problem}
This section formalizes the cooperative construction task and its private-information structure. We first define the workspace, actions, and feasibility constraints, and then formulate the interaction as a Dec-POMDP that motivates the proposed computational architecture.

\subsection{La Boca Construction Task}\label{sec:laboca-task}
\begin{figure}[hb!]
\centering
\includegraphics[width=0.5\linewidth]{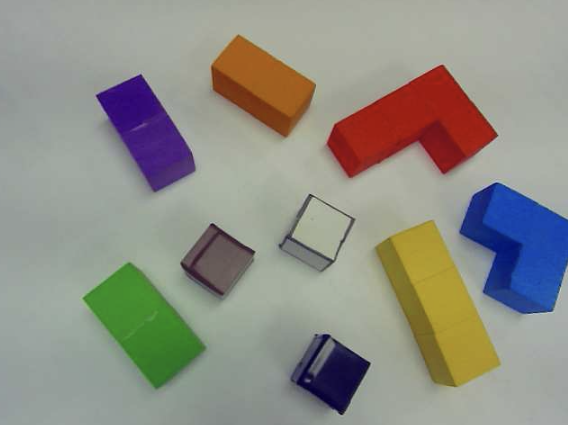}
\caption{The nine construction pieces used in the experiment. The pieces differ in shape and color: white, black, brown, purple, green, orange, yellow, blue, and red.}
\label{fig:pieces}
\end{figure}

We consider a cooperative construction task inspired by the La Boca game.
The workspace is the discrete grid $ \mc W=\{1,2,3,4\}^{3}$, 
and the team is provided with a finite set of rigid, colored pieces \(\mc P\) shown in Fig.~\ref{fig:pieces}.
 
Let $O_t:\mc W\rightarrow \mc P\cup\{\emptyset\}$
denote the workspace occupancy at decision step \(t\), and let
\(R_t\subseteq\mc P\) denote the set of pieces that have not yet been placed.
We define $x_t=(O_t,R_t)$
as the physical state of the construction.
The human and robot receive private two-dimensional target views $ g^H\in\mc G^H$ and $g^R\in\mc G^R$, respectively, 
corresponding to projections from their viewing directions. Let \(\Pi_H\) and \(\Pi_R\) denote the associated projection operators. A final
structure is successful when $\Pi_H(O_T)=g^H$, $\Pi_R(O_T)=g^R$, and $ R_T=\varnothing$.
Thus, the agents need to construct a physically feasible, possibly non-unique, structure that is
simultaneously consistent with both private views.

A physical action by agent \(i\in\{H,R\}\) is represented as $ a_t^i=(\alpha_t^i,p_t^i,c_t^i,o_t^i)$, 
where \(\alpha_t^i\in\{\texttt{place},\texttt{remove},\texttt{wait}\}\) is
the operation, \(p_t^i\) is the selected piece, \(c_t^i\in\mc W\) is an
anchor cell, and \(o_t^i\) is an orientation when applicable. Let $\operatorname{Occ}(p,c,o)\subseteq\mathbb{Z}^{3}$
denote the cells occupied by piece \(p\) when placed at \(c\) with orientation
\(o\). A placement is admissible only if (i) $p\in R_t$, (ii) $\operatorname{Occ}(p,c,o)\subseteq\mc W$, (iii) $O_t(q)=\emptyset,$ for  each $q\in\operatorname{Occ}(p,c,o)$, and (iv) every newly occupied cell above the ground is supported:
\begin{equation*}
    (x,y,z-1)
    \in
    \operatorname{Occ}(p,c,o)
    \cup
    \{q\in\mc W:O_t(q)\neq\emptyset\}
\end{equation*}
for every
\((x,y,z)\in\operatorname{Occ}(p,c,o)\) with \(z>1\).
Removal actions are admissible only for pieces currently present in the
workspace. We denote the resulting set of admissible physical actions by
\(\mc A_{\mathrm{feas}}(x_t)\).

The agents may also exchange natural-language messages. Let
\(m_t^i\in\mc M^i\) denote the message generated by agent \(i\), where a null
message represents no communication. The complete decision of agent \(i\) is
therefore $ u_t^i=(a_t^i,m_t^i)$.
The task is turn-based. Let \(\rho_t\in\{H,R\}\) denote the active player.
Physical actions,
workspace changes, and exchanged messages are observed by both agents.

\subsection{Dec-POMDP Formulation}

We model the task as a
finite-horizon Dec-POMDP with state $ S_t=(x_t,g^H,g^R,\rho_t)\in\mc S$,
where \(x_t\) is the physical construction state, \(g^H\) and \(g^R\) are the
fixed private target views, and \(\rho_t\in \{H,R\}\) identifies the active
player. The physical state and active player are publicly observed, whereas
\(g^H\) and \(g^R\) are observed only by the human and robot, respectively.

At stage \(t\), agent $i\in \{H,R\}$ selects a decision $u_t^i=(a_t^i,m_t^i)\in\mc U^i$,
consisting of a physical action and a message. The inactive agent selects a
null physical action. The state evolves according to a transition kernel $ P(s'\mid s,u^H,u^R)$, 
which is induced by the deterministic construction dynamics and the
turn-taking rule; the private goals remain fixed throughout the trial.

Because the workspace, executed actions, and exchanged messages are publicly
observed, the common interaction history at stage \(t\) is $C_t=
    \left(
        x_{0:t},
        \rho_{0:t},
        u_{0:t-1}^H,
        u_{0:t-1}^R
    \right)$.
The information available to the two agents is therefore $ I_t^H=(C_t,g^H)$ and $I_t^R=(C_t,g^R)$, respectively. 
A decentralized policy is a collection of stochastic decision rules $\pi_t^i:\mc I_t^i\rightarrow\Delta(\mc U^i)$ such that $u_t^i\sim\pi_t^i(\cdot\mid I_t^i)$, 
for \(i\in\mc \{H,R\}\), where \(\Delta(\mc U^i)\) denotes the set of probability distributions over \(\mc U^i\). The corresponding team objective is
\[
    \max_{\pi^H,\pi^R}
    \bb E^\pi
    \left[
        \sum_{t=0}^{T-1}
        r_t(S_t,u_t^H,u_t^R)
        +r_T(S_T)
    \right],
\]
where \(r_T\) rewards successful completion and the stage rewards penalize
interaction time, unnecessary corrective actions, and invalid proposals.

In the human--robot setting, the robot neither observes \(g^H\) nor knows the
human decision rule \(\pi^H\). We therefore retain an approximate belief over
the human's private target,
\[
    b_t^H(g)
    =
    \Pr(g^H=g\mid C_t,g^R), \quad \text{for each }
    g\in\mc G^H (g^R),
\]
where $\mc G^H(g^R)$ is the set of all feasible target views for the human given the robot's target view.

Let \(e_t^H=(a_t^H,m_t^H)\) denote the observable evidence generated during a
human turn. Given a human response model $q_t^H(e\mid g,C_t,g^R)$, Bayes' rule yields
\begin{equation}
    b_{t+1}^H(g)
    =
    \frac{
        q_t^H(e_t^H\mid g,C_t,g^R)b_t^H(g)
    }{
        \displaystyle
        \sum_{\bar g\in\mc G^H}
        q_t^H(e_t^H\mid \bar g,C_t,g^R)b_t^H(\bar g)
    }.
    \label{eq:goal_belief_update}
\end{equation}
Thus, implementing the belief update requires a model of the human's goal-conditioned behavior.

\subsection{Computational and Architectural Requirements}

A foundational approach to Dec-POMDPs is the common-information framework,
which reformulates the decentralized problem as a centralized belief MDP. A
fictitious coordinator maintains a belief over the latent state and agents'
private information and selects prescriptions for the
agents~\cite{nayyar2013decentralized}. Although this yields a dynamic-programming
characterization of an optimal decentralized policy, direct solution is
generally impractical because the belief space is continuous and the
prescription space includes both physical actions and communication. A
tractable architecture must therefore approximate belief-dependent action
selection.

Human--robot interaction introduces an additional challenge: updating the
robot-side belief \(b_t^H\) requires a goal-conditioned model of human
behavior. A standard Bayesian theory-of-mind model may treat the human as a
noisy optimizer, for example through a Boltzmann
policy~\cite{dragan2013legibility,baker2009action}, but evaluating this model may require
solving a separate decision problem for each candidate goal. The architecture
therefore requires tractable surrogates for both human-response modeling and
belief-dependent planning.

Finally, approximate planning may produce suboptimal proposals, but physically
infeasible actions cannot be executed. Action generation must therefore be
separated from deterministic verification of the constraints in
Section~\ref{sec:laboca-task}, with rejected proposals returned for revision.
Because conversation also conveys information about goals and intended actions, it
must inform both inference and planning. These requirements motivate the
structured LLM architecture developed next.

\section{Structured LLM Coordination Architecture}
\label{sec:architecture}

\begin{figure}[ht!]
\centering
\resizebox{\columnwidth}{!}{%
\begin{tikzpicture}[
  node distance=5mm and 6mm
]

\node[
  block,
  rotate=90,
  minimum width=3.0cm,
  minimum height=0.95cm
] (env) {Environment};

\node[
  llmblock,
  right=14mm of env,
  yshift=-4mm,
  minimum width=2.0cm
] (planner) {hierarchical\\LLM planner};

\node[
  exactblock,
  right=10mm of planner,
  minimum width=1.9cm
] (verifier) {action verifier};

\node[
  block,
  right=10mm of verifier,
  minimum width=2.1cm
] (update) {execute action\\update state};

\node[
  llmblock,
  below=12mm of verifier,
  minimum width=2.0cm
] (tom) {LLM-based\\ToM module};

\node[
  block,
  left=14mm of tom,
  minimum width=1.8cm
] (goal) {human-goal\\belief};

\node[
  llmblock,
  right=10mm of tom,
  minimum width=1.9cm
] (conversation) {conversation\\module};

\draw[arr]
  ([yshift=1.1cm]env.south)
  --
  node[flowlabel, above=1mm] {world \\ state}
  node[flowlabel, below=1mm] {robot \\ target}
  (planner.west);

\draw[arr]
  (planner) --
  node[flowlabel, above=1mm] {candidate \\ action}
  (verifier);

\draw[arr]
  (verifier) --
  node[flowlabel, above=1mm] {valid}
  (update);

\draw[darr]
  (verifier.south)
  -- ++(0,-5mm)
  -|
  node[flowlabel, above=1mm, pos=0.25] {failure feedback}
  ([xshift=6mm]planner.south);

\draw[arr]
  (update.east)
  -- ++(2mm,0mm)
  -- ++(0mm,-31mm)
  -| (env.west);

\draw[darr]
  ([xshift=2mm]env.west)
  -- ++(0,-3mm)
  -|
  node[flowlabel, above=1mm, pos=0.25] {observed actions and placements}
  (tom.south);

\draw[arr]
  (conversation) --
  node[flowlabel, above=1mm] {messages}
  (tom);

\draw[arr]
  (tom) --
  node[flowlabel, above=1mm] {belief update}
  (goal);

\draw[arr]
  ([xshift=3.5mm]goal.north)
  -|
  node[flowlabel, above=1mm,rotate=90, pos=0.65] {goal \\ belief}
  (planner.south);

\draw[darr]
  (conversation.north)
  -- ++(0,3mm)
  -- node[flowlabel, above, pos=0.25]
       {plan-relevant information}
     ([xshift=4mm,yshift=-8mm]planner.south)
  -- ([xshift=4mm]planner.south);

  \draw[arr]
  ([yshift=-6mm]conversation.south)
  -- node[xshift=1mm, flowlabel, right,align=center,pos=0.25]
  {human  \\ dialogue}
  (conversation.south);
  
\end{tikzpicture}%
}
\caption{Structured LLM architecture for human--robot coordination under
private information. LLM modules perform ToM inference,
hierarchical planning, conversation processing, and feedback-based replanning,
while a deterministic verifier enforces physical feasibility before
execution.}
\label{fig:architecture}
\end{figure}
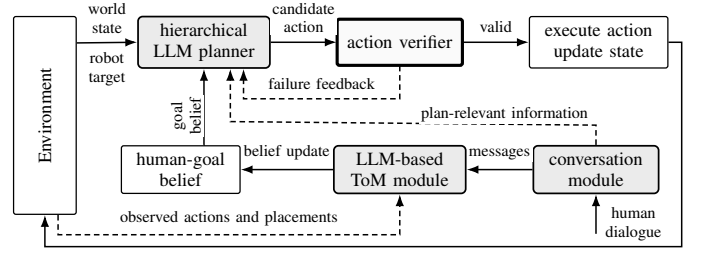

We instantiate the requirements identified in Section~\ref{sec:problem}
through the architecture shown in Fig.~\ref{fig:architecture}. LLM modules
perform ToM inference, hierarchical planning, conversation processing, and
feedback-based replanning, while a rule-based verifier enforces physical
feasibility.

The physical construction state is maintained using an explicit state representation. A structured interaction record stores
previous human and robot actions, selected subgoals, verifier feedback, and task-relevant information extracted from conversation. Together with the robot's
target and current human-goal belief, this record provides the context used by
the LLM modules.

\subsection{Hierarchical LLM Task Planner}
\label{sec:planner}

The task planner reasons about the current construction, the robot's target
view, the remaining pieces, and the observed behavior of the human partner.
Rather than generating a physical action in a single step, it decomposes
decision making into high-level subgoal selection and low-level action
generation.

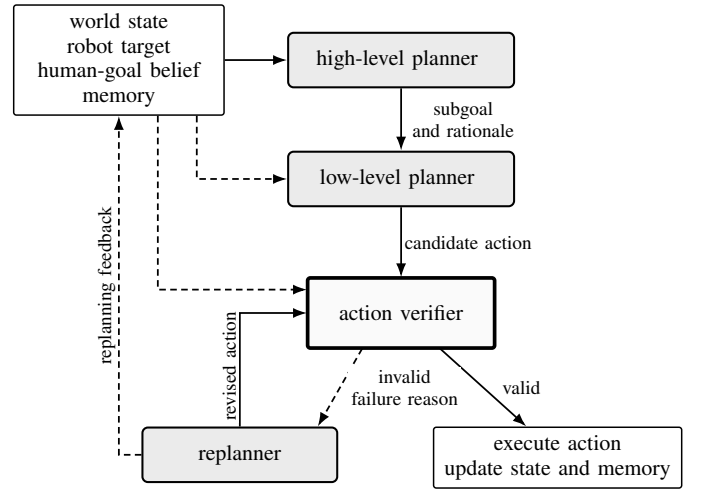
\begin{figure}[ht!]
\centering
\resizebox{\columnwidth}{!}{%
\begin{tikzpicture}[
  node distance=7mm and 14mm
]

\node[
  block,
  minimum width=2.7cm,
  minimum height=1.15cm
] (inputs) {
  world state\\
  robot target\\
  human-goal belief\\
  memory
};

\node[
  llmblock,
  right=8mm of inputs,
  minimum width=2.9cm
] (high) {
  high-level planner
};

\node[
  llmblock,
  below=8mm of high,
  minimum width=2.9cm
] (low) {
  low-level planner
};

\node[
  exactblock,
  below=9mm of low,
  minimum width=2.4cm,
  minimum height=9mm,
  inner sep=1mm
] (verifier) {action verifier};

\node[
  llmblock,
  below left=10mm and -4mm of verifier,
  minimum width=2.5cm
] (replan) {replanner};

\node[
  block,
  below right=10mm and -8mm of verifier,
  minimum width=2.7cm
] (update) {
  execute action\\
  update state and memory
};

\draw[arr]
  (inputs.east) -- (high.west);

\draw[arr]
  (high) --
  node[flowlabel, right=1mm] {subgoal\\and rationale}
  (low);

\draw[arr]
  (low) --
  node[flowlabel, right] {candidate action}
  (verifier);

\draw[arr]
  ([xshift=5mm]verifier.south) --
  node[flowlabel, right=2mm] {valid}
  ([xshift=-4mm]update.north);

\draw[darr]
  ([xshift=-5mm]verifier.south) --
  node[flowlabel, right=1mm] {invalid\\failure reason}
  ([xshift=10mm]replan.north);

\draw[arr]
  (replan.north)
  -- ++(0,13mm)
  node[flowlabel, pos=0.60, above, rotate=90]
       {revised action}
  |- (verifier.west);

\draw[darr]
  (replan.west)
  -|
  node[flowlabel, pos=0.75, above, rotate=90]
       {replanning feedback}
  (inputs.south);

\draw[darr]
  ([xshift=10mm]inputs.south)
  |- (low.west);

\draw[darr]
  ([xshift=5mm]inputs.south)
  |- ([yshift=3mm]verifier.west);

\end{tikzpicture}%
}
\caption{Hierarchical planning with verified replanning. The LLM first
selects a construction subgoal and then generates a candidate action. The
rule-based verifier checks feasibility, and invalid proposals are revised
using structured failure feedback.}
\label{fig:planning-module}
\end{figure}

Let \(\mathcal K_t\) denote the planning context containing the current
workspace state \(x_t\), robot target \(g^R\), human-goal belief \(b_t^H\), and
relevant interaction history. The planning hierarchy is shown in Fig.~\ref{fig:planning-module}, and is summarized by
\[
    z_t
    =
    \mathsf{LLM}_{\mathrm{high}}(\mathcal K_t),
    \quad \text{and} \quad 
    \widetilde a_t^R
    =
    \mathsf{LLM}_{\mathrm{low}}(\mathcal K_t,z_t),
\]
where \(z_t\) is an intermediate construction subgoal and
\(\widetilde a_t^R\) is a candidate physical action.

\paragraph{High-Level Planner:}
The high-level planner identifies the next intermediate objective required to
advance the construction. It compares the current workspace configuration
with the robot's target view while accounting for available pieces, the
current human-goal belief, and prior interaction. Its output contains:
\begin{itemize}
    \item \emph{intent}: the objective of the current planning step;
    \item \emph{subgoal}: a natural-language description of the desired
    intermediate construction state; and
    \item \emph{rationale}: an explanation of how the subgoal contributes to
    task completion.
\end{itemize}
The subgoal provides a persistent objective that may be retained while
individual action proposals are revised.

\paragraph{Low-Level Planner:}
The low-level planner translates the selected subgoal into a candidate action.
Its output follows the action representation introduced in
Section~\ref{sec:laboca-task} and specifies the operation, selected piece, and occupied grid cells.
The candidate may place or
remove a piece, but it is not executed until it passes the
rule-based verifier.

\paragraph{Rule-Based Action Verifier:}
Because LLM-generated actions may violate physical or task constraints, action
generation is separated from feasibility checking. The verifier is implemented
as a rule-based Python function and evaluates
\[
    \mathsf V(x_t,\widetilde a_t^R)
    =
    \begin{cases}
        (\texttt{valid},\varnothing),
        & \widetilde a_t^R\in\mathcal A_{\mathrm{feas}}(x_t),\\
        (\texttt{invalid},e_t),
        & \text{otherwise},
    \end{cases}
    \label{eq:action_verifier}
\]
where \(e_t\) identifies the violated constraint. The verifier checks piece
availability, workspace boundaries, cell overlap, structural support, and
operation-specific rules. It determines whether the action is executable given the current state, not whether it is strategically optimal.

\paragraph{Feedback-Based Replanner:}

When a proposal fails verification, the replanner receives the planning
context, current subgoal, rejected action, and failure explanation. It revises
the operation, piece, placement, or orientation while preserving the high-level
subgoal when appropriate. The revised proposal is then returned to the verifier
before execution.

This verification--replanning loop permits approximate and potentially
suboptimal planning while preventing physically infeasible actions from
reaching the environment. The verifier's structured failure explanation also
provides more informative feedback than a binary rejection.

\subsection{LLM-Based Theory-of-Mind Module}
\label{sec:tom}

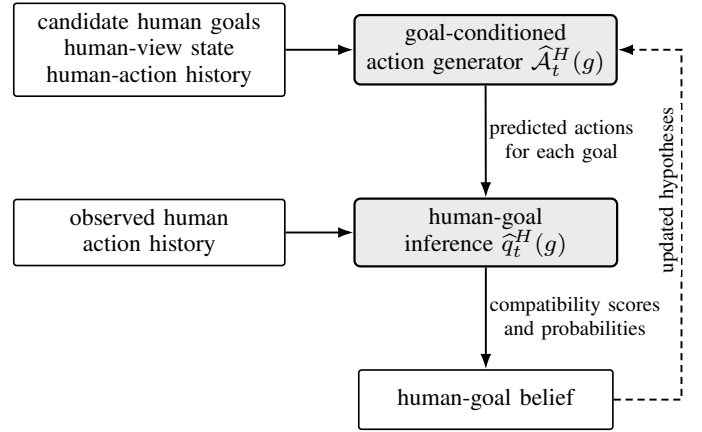
\begin{figure}[ht!]
\centering
\resizebox{\columnwidth}{!}{%
\begin{tikzpicture}[
  node distance=10mm
]

\node[
  block,
  minimum width=3.2cm
] (geninput) {
  candidate human goals\\
  human-view state\\
  human-action history
};

\node[
  block,
  below=12mm of geninput,
  minimum width=3.2cm
] (inferinput) {
  observed human\\
  action history
};

\node[
  llmblock,
  right=8mm of geninput,
  minimum width=3.0cm
] (generator) {
  goal-conditioned\\
  action generator $\widehat{\mathcal A}_t^H(g)$
};

\node[
  llmblock,
  right=8mm of inferinput,
  minimum width=3.1cm
] (inference) {
  human-goal\\
  inference $\widehat q_t^H(g)$
};

\draw[arr]
  (geninput.east) -- (generator.west);

\draw[arr]
  (generator.south) --
  node[flowlabel, right] {predicted actions\\for each goal}
  (inference.north);

\draw[arr]
  (inferinput.east) -- (inference.west);

\node[
  block,
  below=12mm of inference,
  minimum width=3.0cm
] (belief) {human-goal belief};

\draw[arr]
  (inference.south) --
  node[flowlabel, right] {compatibility scores\\and probabilities}
  (belief.north);

\draw[darr]
  (belief.east)
  -- ++(8mm,0)
  |-
  node[flowlabel, pos=0.30, above, rotate=90]
       {updated hypotheses}
  (generator.east);

\end{tikzpicture}%
}
\caption{Action-conditioned Theory-of-Mind inference. For each candidate
human goal, the LLM generates plausible human actions and compares them with
the observed action history to update the robot's belief over the human's
target view.}
\label{fig:tom-module}
\end{figure}

The human and robot observe the target construction from different viewpoints.
The robot must therefore infer the human's private target from observed actions
and communication. The LLM-based Theory-of-Mind (ToM) module, shown in Fig.~\ref{fig:tom-module}, implements the
Bayesian inverse-planning structure underlying the belief update in
\eqref{eq:goal_belief_update}: it predicts human behavior under each candidate
goal and compares these predictions with the behavior actually observed.

For each candidate human target \(g\in\mathcal G^H\), the module computes
\begin{align}
    \widehat{\mathcal A}_t^H(g)
    &=
    \mathsf{LLM}_{\mathrm{act}}
    \bigl(
        x_t^H,g,h_t
    \bigr), \nonumber\\
    \widehat q_t^H(g)
    &=
    \mathsf{LLM}_{\mathrm{compare}}
    \bigl(
        h_t^H,\widehat{\mathcal A}_{0:t}^H(g)
    \bigr), \nonumber
\end{align}
where \(x_t^H\) denotes the current construction represented from the human
viewpoint, \(h_t\) is the interaction history, and \(h_t^H\) is the observed
human-action history. The first module generates goal-conditioned behavioral
predictions, and the second assesses their consistency with the observed
behavior.

The resulting compatibility values instantiate the approximate
human-response likelihood in \eqref{eq:goal_belief_update}. 
The module then outputs a belief distribution representing the probability that the human is pursuing each possible goal.
which is supplied to both the planner and
conversation modules. The belief is revised as additional human placements,
removals, confirmations, and messages become available. This construction
provides an approximate forward model of human behavior without computing an
optimal policy separately for every candidate target.

\subsection{Conversation Module}
\label{sec:conversation}

\begin{figure}[ht!]
\centering
\resizebox{\columnwidth}{!}{%
\begin{tikzpicture}[
  node distance=7mm and 10mm
]

\node[
  block,
  minimum width=2.5cm
] (transcript) {human-speech\\transcript};

\node[
  llmblock,
  below=8mm of transcript,
  minimum width=2.6cm
] (extract) {
  information\\
  extraction
};

\node[
  llmblock,
  below=8mm of extract,
  minimum width=2.4cm
] (response) {robot-response\\generation};

\node[
  exactblock,
  below=8mm of response,
  minimum width=2.6cm
] (decision) {
  reply /\\
  plan-change decision
};

\node[
  block,
  right=16mm of response,
  minimum width=2.3cm
] (memory) {structured\\memory};

\node[
  block,
  below=8mm of decision,
  minimum width=2.3cm
] (update) {plan update};

\node[
  left=12mm of response,
  align=left
] (context) {};

\draw[arr] 
  (transcript) -- (extract);

\draw[arr]
  (extract) --
  node[flowlabel, right=1mm] {structured summary\\human suggestion}
  (response);

\draw[arr]
  (context.east) --
  node[flowlabel, above=1mm] {world state\\robot target}
    node[flowlabel, below=1mm] {human-goal \\ belief}
  (response.west);

\draw[arr]
  (response) -- (decision);

\draw[arr]
  (transcript.east) -|
  node[flowlabel, pos=0.25, above] {transcript}
  (memory.north);

\draw[arr]
  ([yshift=1.2mm]response.east) --
  node[flowlabel, above=1mm] {new interaction}
  ([yshift=1.2mm]memory.west);

\draw[darr]
  ([yshift=-1.2mm]memory.west) --
  node[flowlabel, below=1mm] {conversation \\ history}
  ([yshift=-1.2mm]response.east);

\draw[arr]
  (decision.east) -| (memory.south);

\draw[arr]
  (decision.south) --
  node[flowlabel, pos=0.5, right] {if needed}
  (update.north);

\draw[darr]
  (decision.west)
  -- ++(-12mm,0)
  |- (transcript.west);

\end{tikzpicture}%
}
\caption{Conversation module. Human speech is converted into structured
task-relevant information that may update the interaction memory, produce a
robot response, and trigger revision of the current plan.}
\label{fig:conversation-module}
\end{figure}
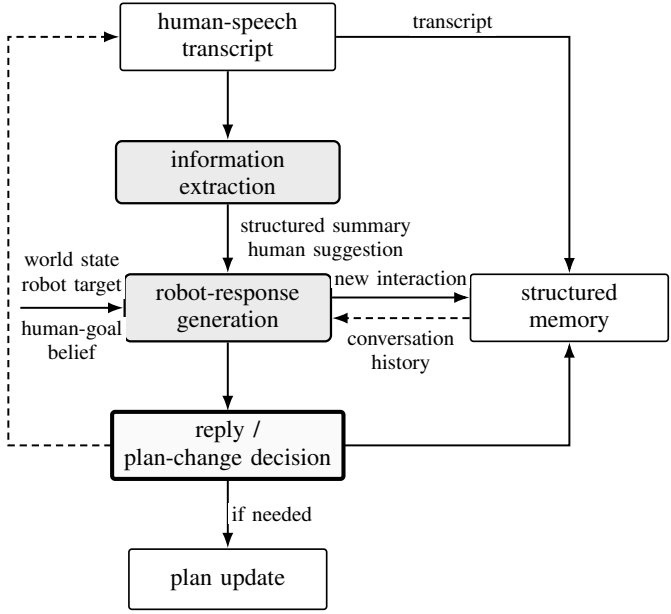

The conversation module, shown in Fig.~\ref{fig:conversation-module}, supports bidirectional natural-language interaction
between the human and robot. It receives the human-speech transcript together
with the current workspace state, robot target, ToM belief, current subgoal,
and relevant conversation history. It extracts task-relevant information and
generates an appropriate robot response.

The robot employs a human speech processing pipeline that converts spoken instructions into structured task information. Human speech is captured through a microphone. The recorded audio is then converted into text using the OpenAI {\small \texttt{gpt-4o-mini-transcribe}} speech recognition model. The resulting transcript is processed by a language understanding module using {\small \texttt{gpt-4.1-nano}} model, which extracts the human’s intent and determines whether the utterance contains an explicit assembly suggestion. If a suggestion is detected, the model converts the natural language instruction into a structured representation containing the action type (place/remove), piece identity, and target coordinates. Finally, the robot’s generated response is converted into speech using a text-to-speech engine to maintain bidirectional spoken communication.
This pipeline allows the human to provide high-level assembly guidance while enabling the robot to interpret, evaluate, and respond to instructions within the task constraints.

The module supports clarification requests, explanations of robot actions,
communication of proposed subgoals and placements, responses to human
questions, negotiation of conflicting proposals, and discussion of alternative
construction strategies. Information extracted from conversation may 
modify the current subgoal, or trigger a new action
proposal.

The human may also provide a specific placement suggestion or broader
strategic guidance. The conversation module interprets the suggestion in a
structured form and evaluates it using the robot's target, the current ToM
belief, and the workspace state. The suggestion may be accepted, modified, or
rejected, and any resulting physical action remains subject to rule-based verification before execution.


\section{Human-Participant Evaluation}
\label{sec:experiments}

We conducted a human-participant study\footnote{The human behavioral experiments were approved under Michigan State University Institutional Review Board Study ID 8969.} to evaluate whether explicit
ToM inference improves coordination efficiency and perceived trust.
We compared the complete structured LLM architecture with an ablation without
the ToM module and with a multi-agent reinforcement-learning policy trained
offline over multiple pairs of private goal views.

\begin{figure}[ht!]
    \centering

    \begin{subfigure}{0.48\linewidth}
        \centering
        \includegraphics[width=\linewidth]{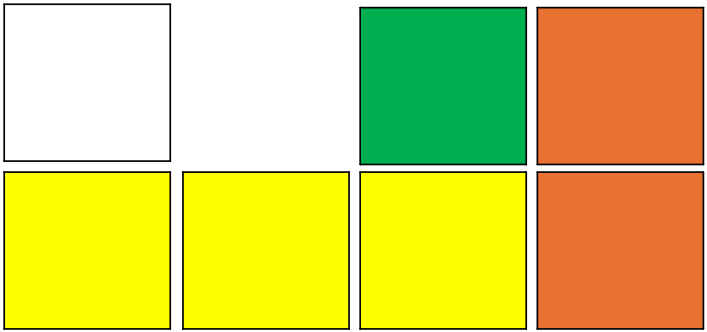}
        \caption{Robot goal}
        \label{fig:robot_goal}
    \end{subfigure}
    \hfill
    \begin{subfigure}{0.48\linewidth}
        \centering
        \includegraphics[width=\linewidth]{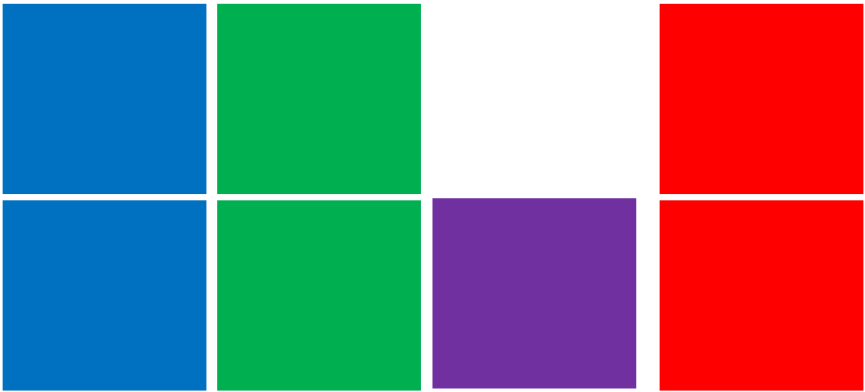}
        \caption{Human goal view 1}
        \label{fig:view1}
    \end{subfigure}

    \vspace{0.3cm}

    \begin{subfigure}{0.48\linewidth}
        \centering
        \includegraphics[width=\linewidth]{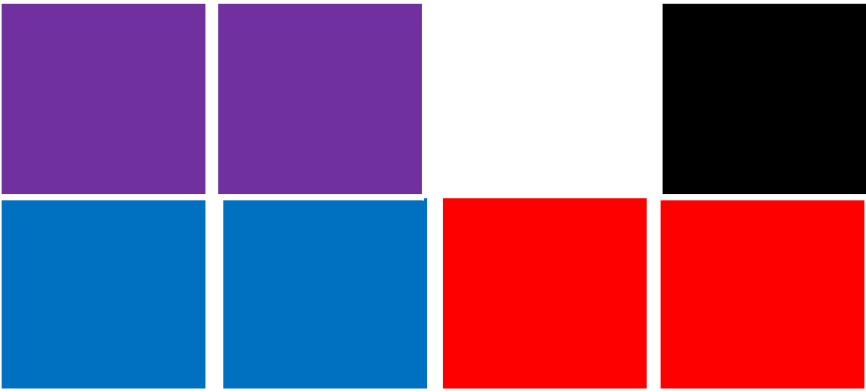}
        \caption{Human goal view 2}
        \label{fig:view2}
    \end{subfigure}
    \hfill
    \begin{subfigure}{0.48\linewidth}
        \centering
        \includegraphics[width=\linewidth]{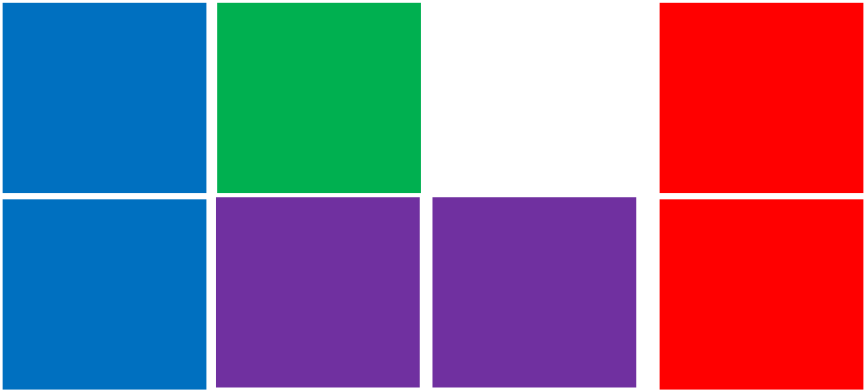}
        \caption{Human goal view 3}
        \label{fig:view3}
    \end{subfigure}

    \caption{Goal-view configurations used in the task: (a) is the target goal assigned to the robot. (b)–(d) are goal views assigned to the human participant, assigned randomly.}
    \label{fig:Goal_views}
\end{figure}

\subsection{Experiment}

\paragraph{Participants and apparatus:}
Five participants were recruited for the pilot study. Each participant
collaborated with a 6-DOF robot to complete the La Boca construction task from
opposing viewpoints. The participant and robot were given distinct private
target views, shown in Fig. \ref{fig:Goal_views}.
Participants were graduate students aged 25–30 years. They communicated with the robot through a wireless speaker equipped with a built-in microphone, which was connected to the computer.

\paragraph{Experimental conditions:}
Each participant completed the task under three conditions. In the
\emph{LLM+ToM} condition, the robot used the complete architecture, including
action-conditioned inference of the human's private target. In the
\emph{LLM} condition, the robot used the same hierarchical planning,
conversation, verification, and replanning modules, but did not maintain an
explicit estimate of the human's target. In the \emph{RL} condition, the robot
used an offline-trained multi-agent RL policy. Goal views for the human were randomly
assigned across trials.
The order in which participants interacted with the policies was randomized at the beginning of the experiment, along with the assignment of the corresponding goal views.

\paragraph{Offline multi-agent RL baseline:}
The RL baseline consists of two MaskablePPO policies trained through
alternating cooperative self-play~\cite{silver2018general,heinrich2015fictitious}. During each training phase, one policy is
updated while the other is held fixed and acts as its partner. Both agents'
actions are applied to the shared construction, and training uses a common
reward based on progress toward both target views.

At execution, each actor observes the current three-dimensional workspace,
the set of unplaced pieces, its current view of the construction, and its
private target view; the partner's target is not available to the actor.
During training, each MaskablePPO policy uses a centralized value-function
critic that additionally observes the partner's current view and private
target. Let \(o_t^i\) denote agent \(i\)'s local observation, and let
\(\bar o_t^i\) denote the augmented observation available to its centralized
critic during training. The actor selects actions $a_t^i$ according to policy $\pi^i(\cdot\mid o_t^i)$, 
while the centralized critic evaluates \(V^i(\bar o_t^i)\). The augmented
observation is used only during training, while execution remains decentralized.

\paragraph{LLM implementation:}
The LLM-based components used GPT-5.4 nano, selected to limit interaction
latency and computational cost. The full and ablated LLM conditions used the
same hierarchical planner, conversation module, action verifier, and
feedback-based replanner; they differed only in whether the ToM estimate was
provided to planning and conversation.

\paragraph{Procedure and measures:}
Participants first completed the Trust Perception Scale--HRI
~\cite{schaefer2016measuring}. They then interacted with each robot policy and
completed the same trust questionnaire after each condition. The principal
task measures were successful completion, participant-initiated termination,
number of interaction steps, completion time, invalid action proposals, and
replanning attempts.

\subsection{Results}

\begin{figure}[ht!]
    \centering
    \begin{subfigure}{0.48\textwidth}
        \centering
        \includegraphics[width=\linewidth]{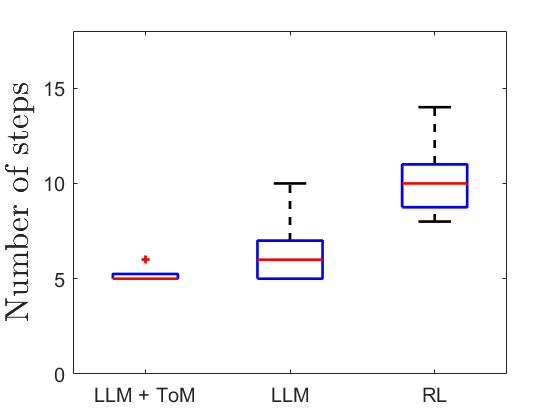}
        \caption{Interaction steps.}
        \label{fig:laboca-steps}
    \end{subfigure}

\vspace{1mm}

    \begin{subfigure}{0.48\textwidth}
        \centering
        \includegraphics[width=\linewidth]{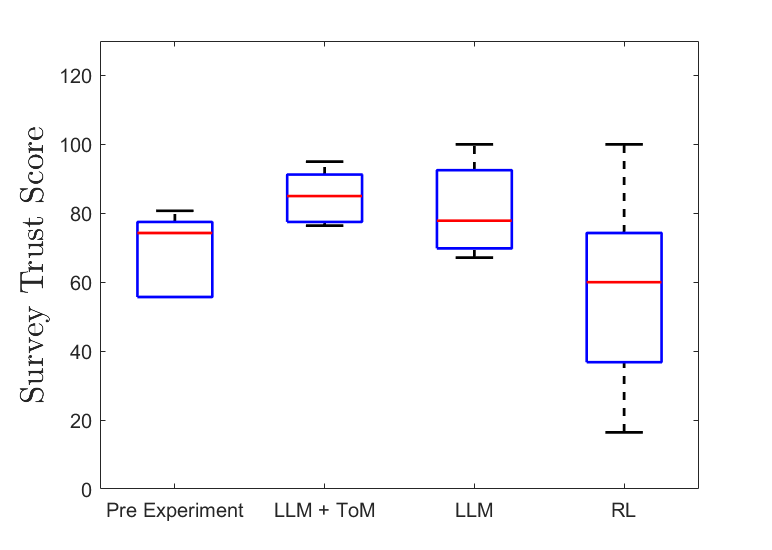}
        \caption{Trust scores.}
        \label{fig:laboca-trust}
    \end{subfigure}

    \caption{Pilot human-participant results. 
    \subref{fig:laboca-steps} Logged interaction steps under the complete
    LLM+ToM architecture, the LLM ablation, and the offline-trained RL policy.
    RL trajectories include participant-terminated trials and therefore should
    not be interpreted as conventional completion times.
    \subref{fig:laboca-trust} Trust Perception Scale--HRI scores collected
    before the experiment and after each policy condition. Individual
    participant observations are overlaid on the box plots.}
    \label{fig:human-study-results}
\end{figure}

\paragraph{Task efficiency and completion:}
As shown in Fig.~\ref{fig:human-study-results}(a), the mean number of logged
interaction steps was \(5.2\) for LLM+ToM, \(6.4\) for the LLM ablation, and
\(10.2\) for RL. The complete architecture therefore required the fewest
steps in this pilot, followed by the architecture without partner-goal
inference.  Using the LLM planners, all participants successfully completed the task.

Only one participant completed the RL condition, doing so in nine steps. The
remaining four participants terminated the interaction after eight to fourteen
steps because of frustration with the robot's behavior. The RL distribution
therefore combines one completed trial with four partial trajectories. The
difference between the two LLM conditions is consistent with the hypothesis
that explicit ToM inference improves coordination.

\paragraph{Trust:}
Figure~\ref{fig:human-study-results}(b) compares the trust scores across
conditions. The mean pre-experiment score was \(68.6\). Mean post-interaction
trust increased to \(84.9\) under LLM+ToM and \(81.1\) under the LLM
ablation, whereas it decreased to \(57.1\) after interaction with the RL
policy. The complete architecture received the highest mean trust score,
although the difference between the two LLM conditions was modest. 

\paragraph{Theory-of-Mind inference:}
To assess the ToM mechanism directly, we examine the probability assigned to
the true human goal after successive observed actions, the number of
observations required for the true goal to become the highest-probability
hypothesis, and changes in the belief induced by conversation.



\begin{figure}[ht!]
    \centering
    \includegraphics[width=0.75\linewidth]{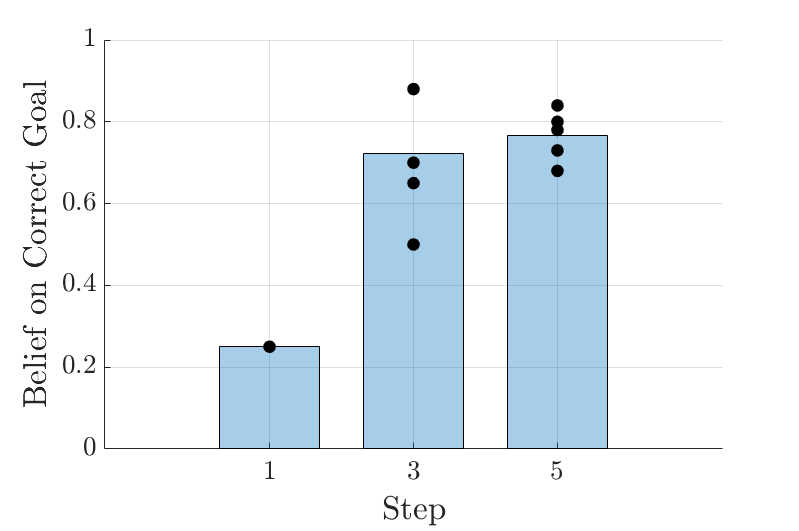}
    \caption{Evolution of belief in the target goal view. The belief assigned to the correct goal view is shown across steps and trials involving human participants with LLM + ToM. The ToM module was executed every two trials; because participants completed the task in an average of 5.2 trials, only two belief updates occurred, resulting in three plotted belief states.}
    \label{fig:avg_bel}
\end{figure}

There are three possible goal views that can be assigned to human participants, along with an additional unassigned view that is not revealed to participants and is included to further evaluate the effectiveness of the ToM module. Therefore, the initial belief distribution over all four possible goal views is uniform, with each goal receiving a probability of 0.25. The average belief assigned to the correct goal view across all participants is shown per ToM inference step in Fig.~\ref{fig:avg_bel}. The belief values are computed across all experiments involving human participants interacting with the LLM + ToM framework. Since the ToM module is executed once every two trials and participants completed the task in an average of 5.2 trials, only two belief updates were performed during each task, resulting in three recorded belief states (initial belief and two subsequent updates). As shown in Fig.~\ref{fig:avg_bel}, the average belief assigned to the correct goal view increases after each ToM update, indicating that the ToM module is able to progressively infer the participant's intended goal.


\subsection{Simulation Results}
In addition to experiments with human participants, we evaluate the performance of different agent configurations in simulation by assigning policies to the roles of each player. Specifically, we compare three scenarios: RL/RL, where both players are controlled by reinforcement learning policies; LLM/RL, where one player is controlled by an LLM-based planner and the other by an RL policy; and LLM/LLM, where both players are controlled by LLM-based planners. These simulations allow us to evaluate the effectiveness of each approach under controlled conditions and compare task success and efficiency across different agent pairings. The simulations are restricted to a maximum of 15 steps, where an episode is considered a failure if the task cannot be completed within this limit. For each configuration, 10 episodes are run, and the results are shown in Table~\ref{tab:simulation_results}.

\begin{table}[ht!]
\centering
\caption{Simulation performance under different agent configurations.}
\label{tab:simulation_results}
\begin{tabular}{lcc}
\hline
\textbf{Configuration} & \textbf{Success Rate (\%)} & \textbf{Avg. Steps} \\
\hline
RL/RL   & 100\% & 5 \\
LLM/RL  & 0\% & 15 \\
LLM/LLM  & 60\% & 12 \\
\hline
\end{tabular}
\end{table}

\section{Discussion and Conclusion}
\label{sec:discussion}

The Dec-POMDP formulation guides a systematic decomposition of team decision
making into ToM inference, hierarchical planning, conversation interpretation,
action verification, and feedback-based replanning. Within this decomposition,
LLMs serve as tractable surrogates for difficult inference and planning
computations, while the rule-based verifier isolates physical feasibility from
semantic reasoning. This separation allows candidate plans to be generated and
revised without relying on the LLM to represent the exact admissible action
set.

The proposed architecture also represents the human-goal estimate, current
subgoal, verification outcome, and plan revision explicitly. This supports
adaptation within an interaction, direct use of natural-language information,
and post hoc interpretation of robot decisions. The human study further indicates that feasibility alone does not ensure effective teaming:
despite action masking, the RL condition frequently produced behavior that
participants found difficult to interpret and untrustworthy.
Overall, the results suggest that a decision-theoretically structured LLM
architecture can support effective human--robot coordination under private
information, improving both task efficiency and the human experience.

Future work will extend the framework to larger workspaces, richer action spaces, and more complex cooperative tasks. Additional directions include calibrated uncertainty estimates for ToM inference, confidence-triggered
clarification, adaptation to unfamiliar partner strategies, and hybrid architectures that combine learned policies with LLM-based reasoning and rule-based verification.

\bibliographystyle{IEEEtran}
\bibliography{references}

\end{document}